\documentclass{article}

\usepackage{amssymb}
\usepackage{graphicx}
\usepackage[caption=false]{subfig}

\usepackage{url}
\newcommand\keywordname{{\bfseries keyword}}
\newcommand\keywords[1]{\par\addvspace\baselineskip
\noindent\keywordname\enspace\ignorespaces#1}

\usepackage{mathtools}
\usepackage{amsfonts}
\usepackage{amsthm}
\usepackage{tabularx}
\usepackage{mdl}

\usepackage[numbers]{natbib}
\usepackage{hyperref}
\usepackage{cleveref}
\usepackage{autonum}

\usepackage{algorithm}
\usepackage{algorithmic}

\newcommand{\p}[1]{{\rm Pr}(#1)}
\renewcommand{\vec}[1]{{\boldsymbol #1}}
\newcommand{\mat}[1]{{\boldsymbol #1}}
\newcommand{\set}[1]{#1}
\newcommand{\fset}[1]{{\mathcal #1}}
\newcommand{\family}[1]{{\mathcal #1}}

\newcommand{\Mean}[2][]{{\rm E}_{#1}\left[ #2 \right]}

\newcommand{\sgn}[1]{{\rm sgn}(#1)}

\DeclareMathOperator*{\argmin}{arg\,min}
\DeclareMathOperator*{\argmax}{arg\,max}

\DeclarePairedDelimiterX\inner[2]{\langle}{\rangle}{#1,#2}
\DeclarePairedDelimiter\norm{\lVert}{\rVert}

\DeclarePairedDelimiter\abs{\lvert}{\rvert}

\newcommand{\RealSet}{\mathbb{R}}

\newtheorem{theorem}{Theorem}

\newtheorem{corollary}{Corollary}
\newtheorem{definition}{Definition}
\newtheorem{proposition}{Proposition}

\crefname{table}{Table}{Tables}
\crefname{figure}{Fig}{Figs}
\crefname{section}{Section}{Sections}
\crefname{equation}{Eq}{Eqs}
\crefname{theorem}{Theorem}{Theorems}
\crefname{definition}{Definition}{Definitions}
\crefname{corollary}{Corollary}{Corollaries}
\crefname{lemma}{Lemma}{Lemmas}
\crefname{proposition}{Proposition}{Propositions}

\begin{document}

\title{Neutralized Empirical Risk Minimization with Generalization Neutrality Bound}


%
%
\author{Kazuto Fukuchi \and Jun Sakuma}
\institute{University of Tsukuba, 1-1-1 Tennodai, Tsukuba, Ibaraki, 305-8577 Japan \\
\email{kazuto@mdl.cs.tsukuba.ac.jp} and \email{jun@cs.tsukuba.ac.jp}
}

%
%

\maketitle

\begin{abstract}
Currently, machine learning plays an important role in the lives and individual activities of numerous people. Accordingly, it has become necessary to design machine learning algorithms to ensure that discrimination, biased views, or unfair treatment do not result from decision making or predictions made via machine learning. In this work, we introduce a novel empirical risk minimization (ERM) framework for supervised learning, neutralized ERM (NERM) that ensures that any classifiers obtained can be guaranteed to be neutral with respect to a viewpoint hypothesis. More specifically, given a viewpoint hypothesis, NERM works to find a target hypothesis that minimizes the empirical risk while simultaneously identifying a target hypothesis that is neutral to the viewpoint hypothesis. Within the NERM framework, we derive a theoretical bound on empirical and generalization neutrality risks. Furthermore, as a realization of NERM with linear classification, we derive a max-margin algorithm, neutral support vector machine (SVM). Experimental results show that our neutral SVM shows improved classification performance in real datasets without sacrificing the neutrality guarantee.
\keywords{neutrality, discrimination, fairness, classification, empirical risk minimization, support vector machine}
\end{abstract}

\section{Introduction}

Within the framework of empirical risk minimization (ERM), a supervised learning algorithm seeks to identify a hypothesis $f$ that minimizes empirical risk with respect to given pairs of {\em input} $x$ and {\em target} $y$. Given an input $x$ without the target value, hypothesis $f$ provides a prediction for the target of $x$ as $y=f(x)$.
In this study, we add a new element, {\em viewpoint hypothesis} $g$, to the ERM framework. Similar to hypothesis $f$, which is given an input $x$ without the viewpoint value, viewpoint hypothesis $g$ provides a prediction for the viewpoint of the $x$ as $v=g(x)$. In order to distinguish between the two different hypotheses, $f$ and $g$, $f$ will be referred to as the {\em target hypothesis}.
Examples of the viewpoint hypothesis are given with the following specific applications.

With this setup in mind, we introduce our novel framework for supervised learning, {\em neutralized ERM} (NERM). Intuitively, we say that a target hypothesis is neutral to a given viewpoint hypothesis if there is low correlation between the target $f(x)$ and viewpoint $g(x)$. The objective of NERM is to find a target hypothesis $f$ that minimizes empirical risks while simultaneously remaining neutral to the viewpoint hypothesis $g$. The following two application scenarios motivate NERM.

{\bf Application 1 (Filter bubble)}
Suppose an article recommendation service provides personalized article distribution. In this situation, by taking a user's access logs and profile as input $x$, the service then predicts that user's preference with respect to articles using supervised learning as $y=f(x)$ (target hypothesis). Now, suppose a user strongly supports a policy that polarizes public opinion (such as nuclear power generation or public medical insurance). Furthermore, suppose the user's opinion either for or against the particular policy can be precisely predicted by $v=g(x)$ (viewpoint hypothesis).
Such a viewpoint hypothesis can be readily learned by means of supervised learning, given users' access logs and profiles labeled with the parties that the users support.
In such situations, if predictions by the target hypothesis $f$ and viewpoint hypothesis $g$ are closely correlated, recommended articles are mostly dominated by articles supportive of the policy, which may motivate the user to adopt a biased view of the policy \cite{resnick2011}. This problem is referred to as the {\em filter bubble} \cite{pariser2011filter}. Bias of this nature can be avoided by training the target hypothesis so that the predicted target is independent of the predicted viewpoint.

{\bf Application 2 (Anti-discrimination)}
Now, suppose a company wants to make hiring decisions using information collected from job applicants, such as age, place of residence, and work experience. While taking such information as input $x$ toward the hiring decision, the company also wishes to predict the potential work performance of job applicants via supervised learning, as $y=f(x)$ (target hypothesis). Now, although the company does not collect applicant information on sensitive attributes such as race, ethnicity, or gender, suppose such sensitive attributes can be sufficiently precisely predicted from an analysis of the non-sensitive applicant attributes, such as place of residence or work experience, as $v=g(x)$ (viewpoint hypothesis). Again, such a viewpoint hypothesis can be readily learned by means of supervised learning by collecting moderate number of labeled examples.
In such situations, if hiring decisions are made by the target hypothesis $f$ and if there is a high correlation with the sensitive attribute predictions $v=g(x)$, those decisions might be deemed discriminatory \cite{pedreschi2009measuring}. In order to avoid this, the target hypothesis should be trained so that the decisions made by $f$ are not highly dependent on the sensitive attributes predicted by $g$. Thus, this problem can also be interpreted as an instance of NERM.

The neutrality of a target hypothesis should not only be guaranteed for given samples, but also for unseen samples. In the article recommendation example, the recommendation system is trained using the user's past article preferences, whereas recommendation neutralization is needed for unread articles. In the hiring decision example, the target hypothesis is trained with information collected from the past histories of job applicants, but the removal of discrimination from hiring decisions is the desired objective.

Given a viewpoint hypothesis, we evaluate the degree of neutrality of a target hypothesis with respect to given and unseen samples as {\em empirical neutrality risk} and {\em generalization neutrality risk}, respectively. The goal of NERM is to show that the generalization risk is theoretically bounded in the same manner as the standard ERM \cite{DBLP:journals/jmlr/BartlettM02,bartlett2005,DBLP:conf/nips/KakadeST08}, and, simultaneously, to show that the generalization neutrality risk is also bounded with respect to given viewpoint hypothesis.

{\bf Our Contribution.}
The contribution of this study is three-fold. First, we introduce our novel NERM framework in which, assuming the target hypothesis and viewpoint hypothesis output binary predictions, it is possible to learn a target hypothesis that minimizes empirical and empirically neutral risks. Given samples and a viewpoint hypothesis, NERM is formulated as a convex optimization problem where the objective function is the linear combination of two terms, the empirical risk term penalizing the target hypothesis prediction error and the neutralization term penalizing correlation between the target and the viewpoint. The predictive performance and neutralization can be balanced by adjusting a parameter, referred to as the neutralization parameter. Because of its convexity, the optimality of the resultant target hypothesis is guaranteed (in \cref{sec:problem}).

Second, we derive a bound on empirical and generalization neutrality risks for NERM. We also show that the bound on the generalization neutrality risk can be controlled by the neutralization parameter (in \cref{sec:generalization-bound}). As discussed in \cref{sec:related}, a number of diverse algorithms targeting the neutralization of supervised classifications have been presented. However, none of these have given theoretical guarantees on generalization neutrality risk. To the best of our knowledge, this is the first study that gives a bound on generalization neutrality risk.

Third, we present a specific NERM learning algorithm for neutralized linear classification. The derived learning algorithm is interpreted as a {\em support vector machine}  (SVM) \cite{vapnik1998statistical} variant with a neutralization guarantee. The kernelized version of the neutralization SVM is also derived from the dual problem (in \cref{sec:neutral-svm}).

\section{Related Works}\label{sec:related}
Within the context of removing discrimination from classifiers, the need for a neutralization guarantee has already been extensively studied. Calders \& Verwer~\cite{Calders:2010:TNB:1842547.1842562} pointed out that elimination of sensitive attributes from training samples does not help to remove discrimination from the resultant classifiers. In the hiring decision example, even if we assume that a target hypothesis is trained with samples that have no race or ethnicity attributes, hiring decisions may indirectly correlate with race or ethnicity through addresses if there is a high correlation between an individual's address and his or her race or ethnicity. This indirect effect is referred to as a {\em red-lining effect} \cite{DBLP:conf/icdm/CaldersKP09}.

Calders \& Verwer~\cite{Calders:2010:TNB:1842547.1842562} proposed the Calders--Verwer 2 Na\"{i}ve Bayes method (CV2NB) to remove the red-lining effect from the Na\"{i}ve Bayes classifier. The CV2NB method is used to evaluate the Calders--Verwer (CV) score, which is a measure that evaluates discrimination of na\"{i}ve Bayes classifiers. The CV2NB method learns the na\"{i}ve Bayes classifier in a way that ensures the CV score is made as small as possible. Based on this idea, various situations where discrimination can occur have been discussed in other studies \cite{zliobaite2011handling,kamiran2010discrimination}. Since a CV score is empirically measured with the given samples, na\"{i}ve Bayes classifiers with low CV scores result in less discrimination for those samples. However, less discrimination is not necessarily guaranteed for unseen samples. Furthermore, the CV2NB method is designed specifically for the na\"{i}ve Bayes model and does not provide a general framework for anti-discrimination learning.

Zemel et al.~\cite{DBLP:conf/icml/ZemelWSPD13} introduced the learning fair representations (LFR) model for preserving classification fairness. LFR is designed to provide a map, from inputs to prototypes, that guarantees the classifiers that are learned with the prototypes will be fair from the standpoint of statistical parity.
Kamishima et al.~\cite{kamishima.ecml.pkdd2012.fairness-aware} presented a prejudice remover regularizer (PR) for fairness-aware classification that is formulated as an optimization problem in which the objective function contains the loss term and the regularization term that penalizes mutual information between the classification output and the given sensitive attributes.
The classifiers learned with LFR or PR are empirically neutral (i.e., fair or less discriminatory) in the sense of statistical parity or mutual information, respectively. However, no theoretical guarantees related to neutrality for unseen samples have been established for these methods.

Fukuchi et al.~\cite{DBLP:conf/pkdd/FukuchiSK13} introduced {\em $\eta$-neutrality}, a framework for neutralization of probability models with respect to a given viewpoint random variable. Their framework is based on maximum likelihood estimation and neutralization is achieved by maximizing likelihood estimation while setting constraints to enforce $\eta$-neutrality. Since $\eta$-neutrality is measured using the probability model of the viewpoint random variable, the classifier satisfying $\eta$-neutrality is expected to preserve neutrality for unseen samples. However, this method also fails to provide a theoretical guarantee for generalization neutrality.

LFR, PR, and $\eta$-neutrality incorporate a hypothesis neutrality measure into the objective function in the form of a regularization term or constraint; however, these are all non-convex. One of the reasons why generalization neutrality is not theoretically guaranteed for these methods is the non-convexity of the objective functions. In this study, we introduce a convex surrogate for a neutrality measure in order to provide a theoretical analysis of generalization neutrality.

\section{Empirical Risk Minimization}

Let $\set{X}$ and $\set{Y}$ be an input space and a target space, respectively. We assume $\set{D}_n = \{(x_i,y_i)\}_{i=1}^{n} \in \set{Z}^n ~(\set{Z} = \set{X}\times\set{Y})$ to be a set of i.i.d. samples drawn from an unknown probability measure $\rho$ over $(\set{Z},\family{Z})$. We restrict our attention to binary classification, $\set{Y}= \{-1,1\}$, but our method can be expanded to handle multi-valued classification via a straightforward modification.
Given the i.i.d. samples, the supervised learning objective is to construct a target hypothesis $f: \set{X} \to \RealSet$ where the hypothesis is chosen from a class of measurable functions $f \in \fset{F}$. We assume that classification results are given by $\mbox{sgn} \circ f(x)$, that is, $y = 1$ if $f(x) > 0$; otherwise $y=-1$.
Given a loss function $\ell: \set{Y}\times\RealSet \to \RealSet^+$, the generalization risk is defined by
\begin{align}
R(f) = \int \ell(y, f(x)) d\rho.
\end{align}
Our goal is to find $f^* \in \fset{F}$ that minimizes the generalization risk $R(f)$. In general, $\rho$ is unknown and the generalization risk cannot be directly evaluated. Instead, we minimize the empirical loss with respect to sample set $\set{D}_n$
\begin{align}
 R_n(f) = \frac{1}{n}\sum_{i=1}^n \ell(y_i, f(x_i)).
\end{align}
This is referred to as {\em empirical risk minimization (ERM)}.

In order to avoid overfitting, a regularization term $\Omega: \fset{F}\to\RealSet^+$ is added to the empirical loss by penalizing complex hypotheses. Minimization of the empirical loss with a regularization term is referred to as {\em regularized ERM (RERM)}.

\subsection{Generalization risk bound}

{\em Rademacher Complexity} measures the complexity of a hypothesis class with respect to a probability measure that generates samples. The Rademacher Complexity of class $\fset{F}$ is defined as
\begin{align}
 \mathcal{R}_n(\fset{F}) = \Mean[D_n, \vec{\sigma}]{\sup_{f \in \fset{F}} \frac{1}{n}\sum_{i=1}^n \sigma_i f(x_i)}
\end{align}
where $\vec{\sigma} = (\sigma_1,...,\sigma_n)^T$ are independent random variables such that $\p{\sigma_i=1}= \p{\sigma_i=-1}=1/2$.
Bartlett \& Mendelson~\cite{DBLP:journals/jmlr/BartlettM02} derived a generalization loss bound using the Rademacher complexity as follows:
\begin{theorem}[Bartlett \& Mendelson~\cite{DBLP:journals/jmlr/BartlettM02}]\label{thm:generalization-lipschitz}
Let $\rho$ be a probability measure on $(\set{Z}, \family{Z})$ and let $\fset{F}$ be a set of real-value functions defined on $\set{X}$, with $\sup\{|f(x)| : f \in \fset{F}\}$ finite for all $x \in \set{X}$. Suppose that $\phi: \RealSet \to [0,c]$ satisfies and is Lipschitz continuous with constant $L_\phi$. Then, with probability at least $1-\delta$, every function in $\fset{F}$ satisfies
 \begin{align}
  R(f) \le R_n(f) + 2L_\phi \mathcal{R}_n(\fset{F}) + c\sqrt{\frac{\ln(2/\delta)}{2n}} .
 \end{align}
\end{theorem}

\section{Generalization Neutrality Risk and Empirical Neutrality Risk}\label{sec:problem}

In this section, we introduce the viewpoint hypothesis into the ERM framework and define a new principle of supervised learning, neutralized ERM (NERM), with the notion of {\em generalization neutrality risk}. Convex relaxation of the neutralization measure is also discussed in this section.

\subsection{+1/$-$1 Generalization neutrality risk}

Suppose a measurable function $g: \set{X} \to \RealSet$ is given. The prediction of $g$ is referred to as the {\em viewpoint} and $g$ is referred to as the {\em viewpoint hypothesis}.
We say the target hypothesis $f$ is neutral to the viewpoint hypothesis $g$ if the target predicted by the learned target hypothesis $f$ and the viewpoint predicted by the viewpoint hypothesis $g$ are not mutually correlating.
In our setting, we assume the target hypothesis $f$ and viewpoint hypothesis $g$ to give binary predictions by ${\rm sgn} \circ f$ and ${\rm sgn} \circ g$, respectively. Given a probability measure $\rho$ and a viewpoint hypothesis $g$, the neutrality of the target hypothesis $f$ is defined by the correlation between ${\rm sgn} \circ f$ and ${\rm sgn} \circ g$ over $\rho$.
If $f(x)g(x)>0$ holds for multiple samples, then the classification ${\rm sgn} \circ f$ closely correlates to the viewpoint ${\rm sgn} \circ g$. On the other hand, if $f(x)g(x) \le 0$ holds for multiple samples, then the classification ${\rm sgn} \circ f$ and the viewpoint ${\rm sgn} \circ g$ are inversely correlating. Since we want to suppress both correlations, our neutrality measure is defined as follows:

\begin{definition}[+1/-1 generalization neutrality risk]
Let $f \in \fset{F}$ and $g \in \fset{G}$ be a target hypothesis and viewpoint hypothesis, respectively. Let $\rho$ be a probability measure over $(\set{Z}, \family{Z})$. Then, the {\em +1/-1 generalization neutrality risk} of target hypothesis $f$ with respect to viewpoint hypothesis $g$ over $\rho$ is defined by
\begin{align}
 C_{\rm sgn}(f,g) = \abs*{ \int \sgn{f(x)g(x)} d\rho }.
\end{align}
\end{definition}

When the probability measure $\rho$ cannot be obtained, a +1/$-$1 generalization neutrality risk $C_{\rm sgn}(f,g)$ can be empirically evaluated with respect to the given samples $\set{D}_n$.

\begin{definition}[+1/$-$1 empirical neutrality risk]
\sloppy Suppose that $D_n=\{(x_i,y_i)\}_{i=1}^n \in \set{Z}^n$ is a given sample set. Let $f \in \fset{F}$ and $g \in \fset{G}$ be the target hypothesis and the viewpoint hypothesis, respectively. Then, the {\em +1/$-$1 empirical neutrality risk} of target hypothesis $f$ with respect to viewpoint hypothesis $g$ is defined by
\begin{align}
 C_{n, {\rm sgn}}(f,g) = \frac{1}{n}\abs*{ \sum_{i=1}^n \sgn{f(x_i)g(x_i)} }. \label{eq:empirical-neutrality-pm}
\end{align}
\end{definition}

\subsection{Neutralized empirical risk minimization (NERM)}
With the definition of neutrality risk, a novel framework, the {\em Neutralized Empirical Risk Minimization} (NERM) is introduced. NERM is formulated as minimization of the empirical risk and empirical +1/$-$1 neutrality risk:
\begin{align}
 \min_{f \in \fset{F}} ~ & R_n(f) + \Omega(f)+ \eta C_{n, {\rm sgn}}(f,g). \label{eq:optimization-pm}
\end{align}
where $\eta > 0$ is the neutralization parameter which determines the trade-off ratio between the empirical risk and the empirical neutrality risk.

\subsection{Convex relaxation of +1/$-$1 neutrality risk}
Unfortunately, the optimization problem defined by \cref{eq:optimization-pm} cannot be efficiently solved due to the nonconvexity of \cref{eq:empirical-neutrality-pm}. Therefore, we must first relax the absolute value function of $C_{{\rm sgn}}(f,g)$ into the max function. Then, we introduce a convex surrogate of the sign function, yielding a convex relaxation of the +1/$-$1 neutrality risk.

By letting $I$ be the indicator function, the +1/$-$1 generalization neutrality risk can be decomposed into two terms:
\begin{align}
   C_{\rm sgn}(f,g)
 =& \abs[\Big]{\underbrace{\int I(\sgn{g(x)} = \sgn{f(x)}) d\rho}_{\mbox{\small prob. that $f$ agrees with $g$}}
  - \underbrace{\int I(\sgn{g(x)} \ne \sgn{f(x)}) d\rho}_{\mbox{\small prob. that $f$ disagrees with $g$}}} \\
 \coloneqq&~ \abs{C^+_{\rm sgn}(f,g) - C^-_{\rm sgn}(f,g)} \label{eq:correct-and-incorrect-ratio}
\end{align}
The upper bound of the +1/$-$1 generalization neutrality risk $C_{\rm sgn}(f,g)$ is tight if $C^+_{\rm sgn}(f,g)$ and $C^-_{\rm sgn}(f,g)$ are close.
Thus, the following property is derived.
\begin{proposition}\label{prop:equability-convert-to-max}
Let $C^+_{\rm sgn}(f,g)$ and $C^-_{\rm sgn}(f,g)$ be functions defined in \cref{eq:correct-and-incorrect-ratio}.
For any $\eta \in [0.5, 1]$, if
 \begin{gather}
  C^{\max}_{\rm sgn}(f,g) \coloneqq \max(C^+_{\rm sgn}(f,g), C^-_{\rm sgn}(f,g)) \le \eta,
  \shortintertext{then}
  C_{\rm sgn}(f,g) = \abs{C^+_{\rm sgn}(f,g) - C^-_{\rm sgn}(f,g)} \le 2\eta - 1.
 \end{gather}
\end{proposition}
\cref{prop:equability-convert-to-max} shows that $C^{\max}_{\rm sgn}(f,g)$ can be used as the generalization neutrality risk instead of $C_{\rm sgn}(f,g)$.
Next, we relax the indicator function contained in $C^\pm_{\rm sgn}(f,g)$.
\begin{definition}[relaxed convex generalization neutrality risk]
Let $f \in \fset{F}$ and $g \in \fset{G}$ be a classification hypothesis and viewpoint hypothesis, respectively. Let $\rho$ be a probability measure over $(\set{Z}, \family{Z})$. Let $\psi: \RealSet \to \RealSet^+$ be a convex function and
\begin{gather}
 C^\pm_\psi(f,g) = \int \psi( \pm g(x)f(x)) d\rho . \label{eq:generalization-neutrality}
\intertext{Then, the {\em relaxed convex generalization neutrality risk} of $f$ with respect to $g$ is defined by}
 C_\psi(f,g) = \max(C^+_\psi(f,g), C^-_\psi(f,g)).
\end{gather}
\end{definition}
The empirical evaluation of relaxed convex generalization neutrality risk is defined in a straightforward manner.
\begin{definition}[convex relaxed empirical neutrality risk]
Suppose $D_n=\{(x_i,y_i)\}_{i=1}^n \in \set{Z}^n$ to be a given sample set. Let $f \in \fset{F}$ and $g \in \fset{G}$ be the target hypothesis and the viewpoint hypothesis, respectively. Let $\psi: \RealSet \to \RealSet^+$ be a convex function and
\begin{gather}
 C^\pm_{n, \psi}(f,g) = \frac{1}{n}\sum_{i=1}^n \psi(\pm g(x_i)f(x_i)) .\label{eq:empirical-neutrality}
\intertext{Then, {\em relaxed convex empirical neutrality risk} of $f$ with respect to $g$ is defined by}
 C_{n, \psi}(f,g) = \max(C^+_{n,\psi}(f,g), C^-_{n,\psi}(f,g)).
\end{gather}
\end{definition}
$C^\pm_{n,\psi}(f,g)$ is convex because it is a summation of the convex function $\psi$.
Noting that $\max(f_1(x),f_2(x))$ is convex if $f_1$ and $f_2$ are convex, $C_{n,\psi}(f,g)$ is convex as well.

\subsection{NERM with relaxed convex empirical neutrality risk}\label{sec:relaxed-nerm}

Finally, we derive the convex formulation of NERM with the relaxed convex empirical neutrality risk as follows:
\begin{gather}
\begin{align}
 \min_{f \in \fset{F}} ~ & R_n(f) + \Omega(f)+ \eta C_{n,\psi}(f,g). \label{eq:optimization}
\end{align}
\end{gather}
If the regularized empirical risk is convex, then this is a convex optimization problem.

The neutralization term resembles the regularizer term in the formulation sense. Indeed, the neutralization term is different from the regularizer in that it is dependent on samples. We can interpret the regularizer as a prior structural information of the model parameters, but we cannot interpret the neutralization term in the same way due to its dependency on samples. PR and LFR have similar neutralization terms in the sense of adding the neutrality risk to objective function, and neither can be interpreted as a prior structural information. Instead, the neutralization term can be interpreted as a prior information of {\em data}. The notion of a prior data information is relevant to {\em transfer learning} \cite{DBLP:journals/tkde/PanY10}, which aims to achieve learning dataset information from other datasets. However, further research on the relationships between the neutralization and transfer learning will be left as an area of future work.

\section{Generalization Neutrality Risk Bound}\label{sec:generalization-bound}

In this section, we show theoretical analyses of NERM generalization neutrality risk and generalization risk. First, we derive a probabilistic uniform bound of the generalization neutrality risk for any $f \in \fset{F}$ with respect to the  empirical neutrality risk $C_{n,\psi}(f,g)$ and the Rademacher complexity of $\fset{F}$. Then, we derive a bound on the generalization neutrality risk of the optimal hypothesis.

For convenience, we introduce the following notation. For a hypothesis class $\fset{F}$ and constant $c \in \RealSet$, we denote $-\fset{F}=\{-f: f \in \fset{F}\}$ and $c\fset{F} = \{cf: f \in \fset{F}\}$. For any function $\phi: \RealSet \to \RealSet$, let $\phi \circ \fset{F} = \{\phi \circ f : f \in \fset{F}\}$. Similarly, for any function $g: \set{X} \to \RealSet$, let $g \fset{F} = \{ h : f \in \fset{F}, h(x) = g(x)f(x) ~\forall x \in \set{X}\}$.

\subsection{Uniform bound of generalization neutrality risk}
\label{sec:generalization-uniform-neutrality}

A probabilistic uniform bound on $C_\psi(f,g)$ for any hypothesis $f \in \fset{F}$ is derived as follows.

\begin{theorem}\label{thm:generalization-neutrality-bound}
Let $C_\psi(f,g)$ and $ C_{n,\psi}(f,g)$ be the relaxed convex generalization neutrality risk and the relaxed convex empirical neutrality risk of $f \in {\cal F}$ w.r.t. $g \in {\cal G }$. Suppose that $\psi: \RealSet \to [0,c]$ satisfies and is Lipschitz continuous with constant $L_\psi$. Then, with probability at least $1-\delta$, every function in $\fset{F}$ satisfies
\begin{align}
  C_\psi(f,g) \le C_{n,\psi}(f,g) + 2L_\psi \mathcal{R}_n(g\fset{F}) + c\sqrt{\frac{\ln(2/\delta)}{2n}}.
 \end{align}
\end{theorem}
As proved by \cref{thm:generalization-neutrality-bound}, $C_{\psi}(f,g) - C_{n,\psi}(f,g)$, the approximation error of the generalization neutrality risk is uniformly upper-bounded by the Rademacher complexity of hypothesis classes $g \fset{F}$ and $O(\sqrt{\ln(1/\delta)/n})$, where $\delta$ is the confidence probability and $n$ is the sample size.

\subsection{Generalization neutrality risk bound for NERM optimal hypothesis}
Let $\hat{f} \in \fset{F}$ be the optimal hypothesis of NERM. We derive the bounds on the empirical and generalization neutrality risks achieved by $\hat{f}$ under the following conditions:

\newdimen\savedprevdepth
\setlength{\savedprevdepth}{\prevdepth}
\nointerlineskip
\noindent
\parbox{.92\linewidth}{
\setlength{\prevdepth}{\savedprevdepth}
\begin{enumerate}
 \item Hypothesis class $\fset{F}$ includes a hypothesis $f_0$ s.t. $f_0(x) = 0$ for $\forall x$, and
 \item the regularization term of $f_0$ is $\Omega(f_0)=0$.
\end{enumerate}
\global\savedprevdepth\prevdepth
}
\hbox{(A)}
\par\prevdepth=\savedprevdepth
The conditions are relatively moderate. For example, consider the linear hypothesis $f(\vec{x})= \vec{w}^T \vec{x}$ and $\Omega(f)=\norm{\vec{w}}_2^2$ ($\ell_2^2$ norm) and let $W \subseteq \RealSet^D$ be a class of the linear hypothesis. If $\vec{0} \in W$, the two conditions above are satisfied.
Assuming that $\fset{F}$ satisfies these conditions, the following theorem provides the bound on the generalization neutrality risk.
\begin{theorem}\label{thm:generalization-algorithm-neutrality}
Let $\hat{f}$ be the optimal target hypothesis of NERM, where the viewpoint hypothesis is $g \in \fset{G}$ and the neutralization parameter is $\eta$. Suppose that $\psi: \RealSet \to [0,c]$ satisfies and is Lipschitz continuous with constant $L_\psi$. If conditions (A) are satisfied, then with probability at least $1-\delta$,
 \begin{align}
  C_\psi(\hat{f},g) \le \psi(0) + \phi(0)\frac{1}{\eta} + 2L_\psi \mathcal{R}_n(g\fset{F}) + c\sqrt{\frac{\ln(2/\delta)}{2n}} .
 \end{align}
\end{theorem}
For the proof of \cref{thm:generalization-algorithm-neutrality}, we first derive the upper bound of the empirical neutrality risk of $\hat{f}$.
\begin{corollary}\label{col:optimal-neutrality-bound}
If the conditions (A) are satisfied, then the empirical relaxed convex neutrality risk of $\hat{f}$ is bounded by
 \begin{align}
  C_{n,\psi}(\hat{f},g) \le \psi(0) + \phi(0)\frac{1}{\eta} .
 \end{align}
\end{corollary}
\cref{thm:generalization-algorithm-neutrality} is immediately obtained from \cref{thm:generalization-neutrality-bound} and \cref{col:optimal-neutrality-bound}.

\subsection{Generalization risk bound for NERM}\label{sec:generalization-risk-nerm}

In this section, we compare the generalization risk bound of NERM with that of a regular ERM. \cref{thm:generalization-lipschitz} denotes a uniform bound of the generalization risk. This theorem holds with the hypotheses which are optimal in terms of NERM and ERM. However, the hypotheses which are optimal in terms of NERM and ERM have different empirical risk values. The empirical risk of NERM is greater than that of ERM since NERM has a term that penalizes less neutrality. More precisely, if we let $\bar{f}$ be the optimal hypothesis in term of ERM, we have
\begin{gather}
 R_n(\hat{f}) - R_n(\bar{f}) \ge 0 . \label{eq:empirical-risk-nerm}
\end{gather}
The reason for this is that empirical risk of any other hypothesis is greater than one of $\bar{f}$ since $\bar{f}$ minimizes empirical risk. Furthermore, due to $\hat{f}$ is a minimizer of $R_n(f)+\eta C_{n,\phi}(f,g)$, we have
\begin{gather}
  R_n(\hat{f})+\eta C_{n,\phi}(\hat{f},g)-R_n(\bar{f})-\eta C_{n,\phi}(\bar{f},g) \le 0\\
  R_n(\hat{f}) - R_n(\bar{f}) \le \eta (C_{n,\phi}(\bar{f},g) - C_{n,\phi}(\hat{f},g)) . \label{eq:empirical-risk-and-neutrality}
\end{gather}
Since the left term of this inequality is greater than zero due to \cref{eq:empirical-risk-nerm}, the empirical risk becomes greater if the empirical neutrality risk becomes lower.

\section{Neutral SVM}\label{sec:neutral-svm}

\subsection{Primal problem}

SVMs~\cite{vapnik1998statistical} are a margin-based supervised learning method for binary classification. The algorithm of SVMs can be interpreted as minimization of the empirical risk with regularization term, which follows the RERM principle. In this section, we introduce a SVM variant that follows the NERM principle.

The soft-margin SVM employs the linear classifier $f(\vec{x}) = \vec{w}^T\vec{x} + b$ as the target hypothesis. In the objective function, the hinge loss is used for the loss function, as $\phi(yf(x)) = \max(0, 1-yf(x))$, and the $\ell_2$ norm is used for the regularization term, $\Omega(f)=\lambda\norm{f}_2^2/2n$, where $\lambda > 0$ denotes the regularization parameter.
In our SVM in NERM, referred to as the neutral SVM, the loss function and regularization term are the same as in the soft-margin SVM. For a surrogate function of the neutralization term, the hinge loss $\psi(\pm g(x)f(x))=\max(0,1 \mp g(x)f(x))$ was employed. Any hypothesis can be used for the viewpoint hypothesis. Accordingly, following the NERM principle defined in \cref{eq:optimization}, the neutral SVM is formulated by

\begin{gather}
 \begin{split}
 \min_{\vec{w},b} ~ & \sum_{i=1}^n \max(0,1-y_i (\vec{w}^T\vec{x}_i + b)) + \frac{\lambda}{2} \norm{\vec{w}}_2^2 + \eta C_{n,\psi}(\vec{w},b,g),
 \end{split}\label{eq:optimization-primal}
\shortintertext{\normalsize where}
  C_{n,\psi}(\vec{w},b,g) = \max(C_{n,\psi}^+(\vec{w},b,g), C_{n,\psi}^-(\vec{w},b,g)), \\
  C_{n,\psi}^\pm(\vec{w},b,g) = \sum_{i=1}^n \max(0, 1 \mp g(\vec{x}_i)(\vec{w}^T\vec{x}_i + b)) .
\end{gather}
Since the risk, regularization, and neutralization terms are all convex, the objective function of the neutral SVM is convex. The primal form can be solved by applying the subgradient method \cite{Shor:1985:MMN:3585} to \cref{eq:optimization-primal}.

\subsection{Dual problem and kernelization}

Next, we derive the dual problems of the problem of \cref{eq:optimization-primal}, from which the neutral SVM kernelization is naturally derived. First, we introduce slack variables $\vec{\xi}, \vec{\xi}^\pm$, and $\zeta$ into \cref{eq:optimization-primal} to represent the primal problem:

\begin{align}
 \min_{\substack{\vec{w}, b,\\ \vec{\xi}, \vec{\xi}^\pm, \zeta}} ~ &\sum_{i=1}^n \xi_i + \frac{\lambda}{2}\norm{\vec{w}}_2^2 + \eta \zeta \label{eq:optimization-constraints}\\
 \mbox{\rm sub to} ~  &\sum_{i=1}^n \xi^+_i \le \zeta
                     , \sum_{i=1}^n \xi^-_i \le \zeta ,
                      1 - y_i(\vec{w}^T\vec{x}_i + b) \le \xi_i ,\\
                     & 1 - v_i(\vec{w}^T\vec{x}_i + b) \le \xi^+_i ,
                      1 + v_i(\vec{w}^T\vec{x}_i + b) \le \xi^-_i, \\
                     & \xi_i \ge 0, \xi^+_i \ge 0, \xi^-_i \ge 0, \zeta \ge 0
\end{align}
where slack variables $\xi_i, \xi^+_i $, and $\xi^-_i$ denote measures of the degree of misclassification, correlation, and inverse correlation, respectively. The slack variable $\zeta$, derived from $\max$ function in $C_{n,\psi}(\vec{w},b,g)$, measures the imbalance of the degree of correlation and inverse correlation.
From the Lagrange relaxation of the primal problem \cref{eq:optimization-constraints}, the dual problem is derived as

\begin{align}
 \max_{\vec{\alpha}, \vec{\beta}^\pm} ~ & \lambda\sum_{i=1}^nb_i - \frac{1}{2}\sum_i^n\sum_j^n a_i a_i k(x_i,x_j) \label{eq:optimization-kernel}\\
 \mbox{\rm sub to} ~ & \sum_i^n a_i = 0,
                      0 \le \alpha_i \le 1, 0 \le \beta^\pm_i, \beta^+_i + \beta^-_i \le \eta
\end{align}
where $b_i = \alpha_i+\beta^+_i+\beta^-_i, a_i = \alpha_iy_i + \beta^+_iv_i - \beta^-_iv_i$. As seen in the dual problem, the neutral SVM is naturally kernelized with kernel function $\vec{x}_i^T\vec{x}_j = k(x_i, x_j)$. The derivation of the dual problem and kernelization thereof is described in the supplemental document in detail.
The optimization of \cref{eq:optimization-kernel} is an instance of {\em quadratic programming (QP)} that can be solved by general QP solvers, although it does not scale well with large samples due to its large memory consumption. In the supplemental documentation, we also show the applicability of the well-known {\em sequential minimal optimization} technique to our neutral SVM.

\section{Experiments}\label{sec:experiment}
In this section, we present experimental evaluation of our neutral SVM for synthetic and real datasets. In the experiments with synthetic data, we experimentally evaluate the change of generalization risk and generalization neutrality risk according to the number of samples, in which their relations are described in \cref{thm:generalization-neutrality-bound}. In the experiments for real datasets, we compare our method with CV2NB \cite{Calders:2010:TNB:1842547.1842562}, PR \cite{kamishima.ecml.pkdd2012.fairness-aware} and $\eta$-neutral logistic regression~($\eta$LR for short)~\cite{DBLP:conf/pkdd/FukuchiSK13} in terms of risk and neutrality risk. The CV2NB method learns a na\'{i}ve Bayes model, and then modifies the model parameters so that the resultant CV score approaches zero. The PR and $\eta$LR are based on maximum likelihood estimation of a logistic regression (LR) model. These methods have two parameters, the regularizer parameter $\lambda$, and the neutralization parameter $\eta$. The PR penalizes the objective function of the LR model with mutual information. The $\eta$LR performs maximum likelihood estimation of the LR model while enforcing $\eta$-neutrality as constraints.
The neutralization parameter of neutral SVM and PR balances risk minimization and neutrality maximization. Thus, it can be tuned in the same manner used to determine the regularizer parameter. The neutralization parameter of $\eta$LR determines the region of the hypothesis in which the hypotheses are regarded as neutral. The tuning strategy of the regularizer parameter and neutralization parameter are different in all these methods. We determined the neutralization parameter tuning range of these methods via preliminary experiments.

\subsection{Synthetic dataset}
In order to investigate the change of generalization neutrality risk with sample size $n$, we performed our neutral SVM experiments for a synthetic dataset. First, we constructed the input $\vec{x}_i \in \RealSet^{10}$ with the vector being sampled from the uniform distribution over $[-1,1]^{10}$. The target $y_i$ corresponding to the input $\vec{x}_i$ is generated as $y_i={\rm sgn}(\vec{w}_y^T\vec{x}_i)$ where $\vec{w}_y \in \RealSet^{10}$ is a random vector drawn from the uniform distribution over $[-1,1]^{10}$. Noises are added to labels by inverting the label with probability $1/(1+\exp(-100 |\vec{w}_y^T\vec{x}_i|))$. The inverting label probability is small if the input $\vec{x}_i$ is distant from a plane $\vec{w}_y^T \vec{x} = 0$. The viewpoint $v_i$ corresponding to the input $\vec{x}_i$ is generated as $v_i={\rm sgn}(\vec{w}_v^T\vec{x}_i)$, where the first element of $\vec{w}_v$ is set as $w_{v,1} = w_{y,1}$ and the rest of elements are drawn from the uniform distribution over $[-1,1]^{9}$. Noises are added in the same manner as the target. The equality of the first element of $\vec{w}_y$ and $\vec{w}_v$ leads to correlation between $y_i$ and $v_i$.
Set the regularizer parameter as $\lambda = 0.05n$. The neutralization parameter was varied as $\eta \in \{0.1, 1.0, 10.0\}$. In this situation, we evaluate the approximation error of the generalization risk and the generalization neutrality risk by varying sample size $n$. The approximation error of generalization risk is the difference of the empirical risk between training and test samples, while that of the generalization neutrality risk is the difference of the empirical neutrality risk between training and test samples. Five fold cross-validation was used for evaluation of the approximation error of the empirical risk and empirical neutrality; the average of ten different folds are shown as the results.

\begin{figure}[tb]
 \centering
 \subfloat[risk]{\includegraphics[width=.30\columnwidth]{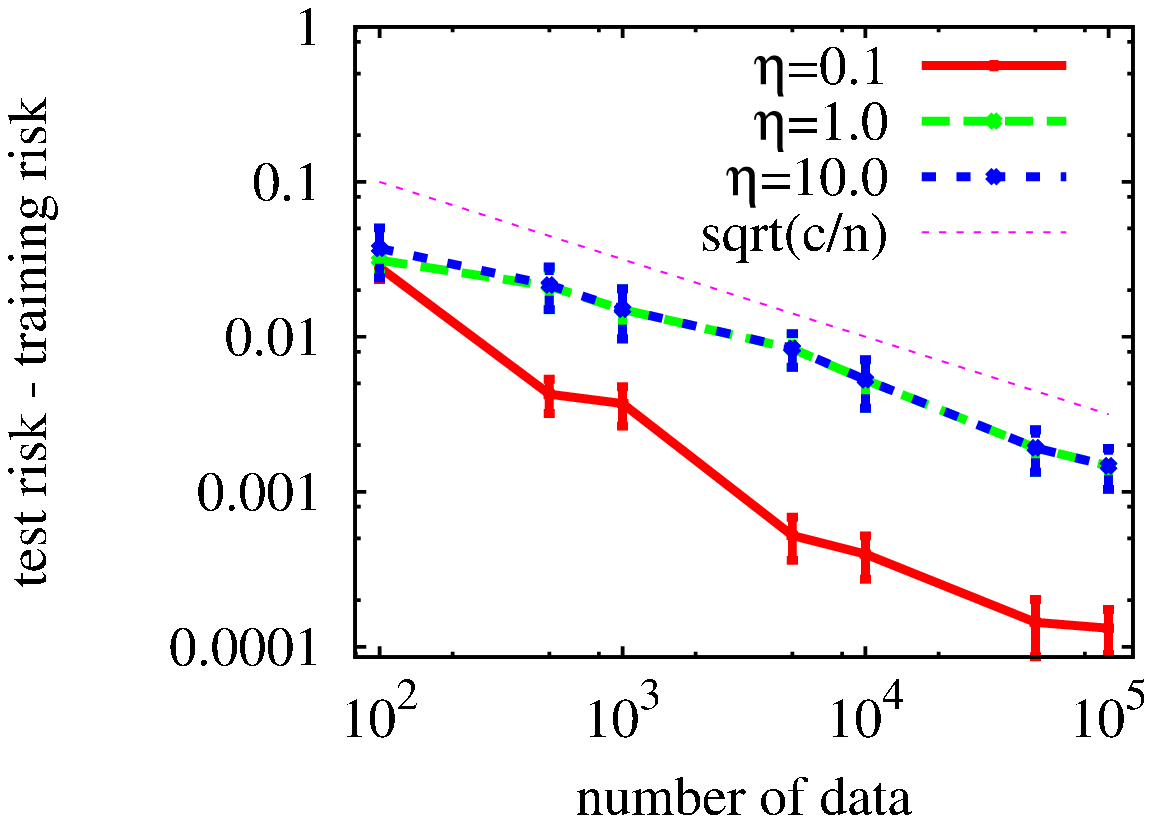}}
 \qquad
 \subfloat[neutrality risk]{\includegraphics[width=.30\columnwidth]{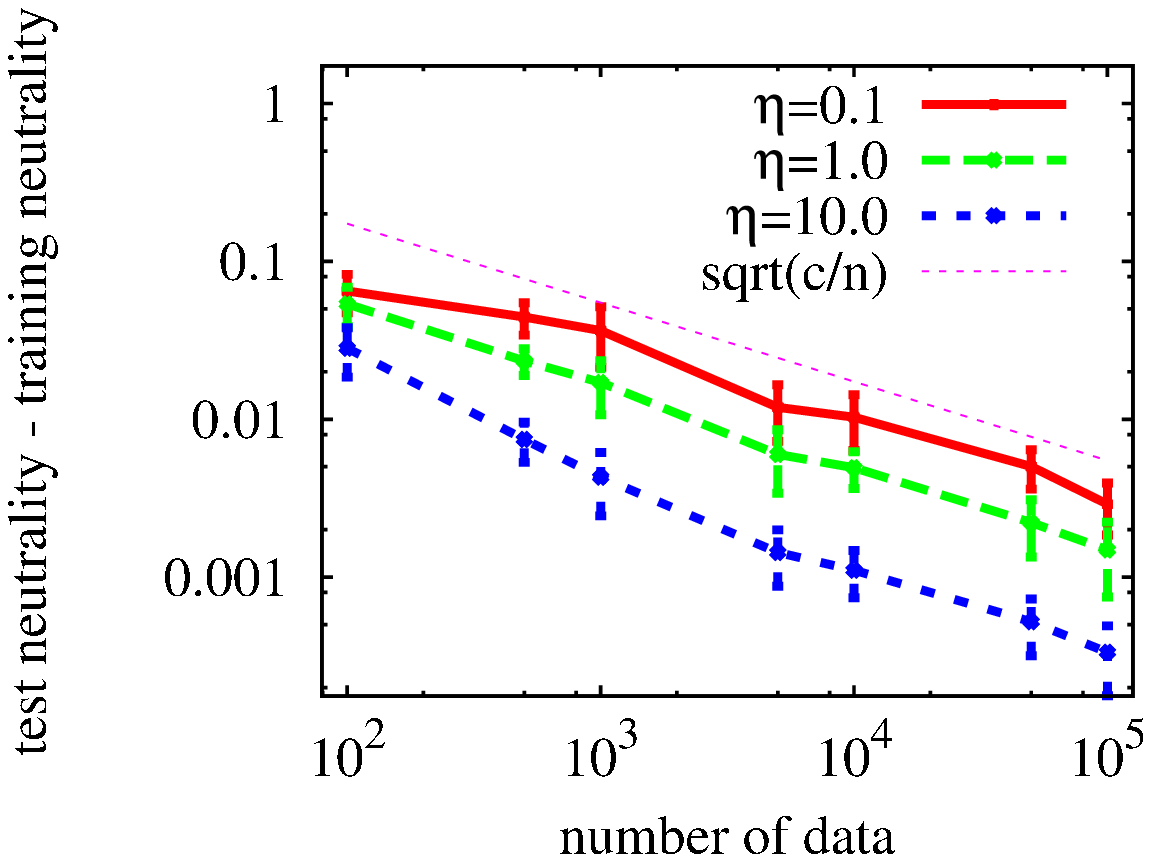}}
 \caption{Change of approximation error of generalization risk (left) and approximation error of generalization neutrality risk (right) by neutral SVM (our proposal) according to varying the number of samples $n$. The horizontal axis shows the number of samples $n$, and the error bar shows the standard deviation across the change of five-fold division. The line ``sqrt(c/n)'' denotes the convergence rate of the approximation error of the generalization risk~(in \cref{thm:generalization-lipschitz}) or the generalization neutrality risk~(in \cref{thm:generalization-neutrality-bound}). Each line indicates the results with the neutralization parameter $\eta \in \{0.1, 1.0, 10.0\}$. The regularizer parameter was set as $\lambda=0.05n$.}\label{fig:synthetic-data-result}
\end{figure}

{\bf Results.}
\cref{fig:synthetic-data-result} shows the change of the approximation error of generalization risk (the difference of the empirical risks w.r.t. test samples and training samples), and the approximation error of generalization neutrality risk (the difference of the empirical neutrality risks w.r.t. test samples and training samples) with changing sample size $n$. The plots in \cref{fig:synthetic-data-result} left and right show the approximation error of generalization risk and the approximation error of generalization neutrality risk, respectively.

Recall that the discussions in \cref{sec:generalization-risk-nerm} showed that the approximation error of generalization risk decreases with $O(\sqrt{\ln(1/\delta)/n})$ rate. As indicated by the \cref{thm:generalization-lipschitz}, \cref{fig:synthetic-data-result} (left) clearly shows that the approximation error of the generalization risk decreases as sample size $n$ increases.
Similarly, discussions in \cref{sec:generalization-uniform-neutrality} revealed that the approximation error of generalization neutrality risk also decreases with $O(\sqrt{\ln(1/\delta)/n})$ rate, which can be experimentally confirmed in \cref{fig:synthetic-data-result} (right). The plot clearly shows that the approximation error of the generalization neutrality risk decreases as the sample size $n$ increases.

\subsection{Real datasets}

\begin{table}[tb]
 \def\tabularxcolumn#1{m{#1}}
 \centering
 \caption{Specification of Datasets}\label{tbl:spec-of-data}
 \newcolumntype{D}{>{\raggedright\arraybackslash\advance\hsize1em}X}
 \newcolumntype{V}{>{\raggedright\arraybackslash\advance\hsize5em}X}
 \newcolumntype{T}{>{\raggedright\arraybackslash\advance\hsize4em}X}
 \newcolumntype{L}{>{\raggedright\arraybackslash}X}
 \newcolumntype{A}{>{\raggedright\arraybackslash\advance\hsize-0.5em}X}
\begin{tabularx}{\columnwidth}{D|LAVT}
\hline
 dataset  & \#Inst.  & \#Attr.  & Viewpoint  & Target  \\
\hline
 Adult   & 16281 & 13 &  gender&   income\\
 Dutch  & 60420 & 10 &  gender&  income\\
 Bank & 45211 & 17 &  loan&  term deposit\\
 German & 1000 & 20 &  foreign worker &  credit risk\\
\hline
\end{tabularx}
\end{table}
\begin{table}[tb]
 \caption{Range of neutralization parameter}\label{tbl:range-of-parameter}
 \begin{tabularx}{\columnwidth}{l|X}
  \hline
   method & range of neutralization parameter \\
  \hline
   PR & 0, 0.01, 0.05, 0.1, ..., 100 \\
   $\eta$LR & 0, $5 \times 10^{-5}$, $1   \times 10^{-4}$, $5 \times 10^{-4}$, ..., 0.5 \\
   neutral SVM & 0, 0.01, 0.05, 0.1, ..., 100 \\
  \hline
 \end{tabularx}
\end{table}

We compare the classification performance and neutralization performance of neutral SVM with CV2NB, PR, and $\eta$LR for a number of real datasets specified in \cref{tbl:spec-of-data}. In \cref{tbl:spec-of-data}, \#Inst. and \#Attr. denote the sample size and the number of attributes, respectively; ``Viewpoint'' and ``Target'' denote the attributes used as the target and the viewpoint, respectively. All dataset attributes were discretized by the same procedure described in \cite{Calders:2010:TNB:1842547.1842562} and coded by 1-of-K representation for PR, $\eta$LR, and neutral SVM. We used the primal problem of neutral SVM (non-kernelized version) to compare our method with the other methods in the same representation.
For PR, $\eta$LR, and neutral SVM, the regularizer parameter was tuned in advance for each dataset in the non-neutralized setting by means of five-fold cross validation, and the tuned parameter was used for the neutralization setting. CV2NB has no regularization parameter to be tuned.
\cref{tbl:range-of-parameter} shows the range of the neutralization parameter used for each method.

The classification performance and neutralization performance was evaluated with {\em Area Under the receiver operating characteristic Curve} (AUC) and +1/$-$1 empirical neutrality risk $C_{n,\rm sgn}(f,g)$, respectively. Both measures were evaluated with five-fold cross-validation and the average of ten different folds are shown in the plots.

\begin{figure}[tb]
 \centering
 \subfloat[Adult]{\includegraphics[width=.32\textwidth]{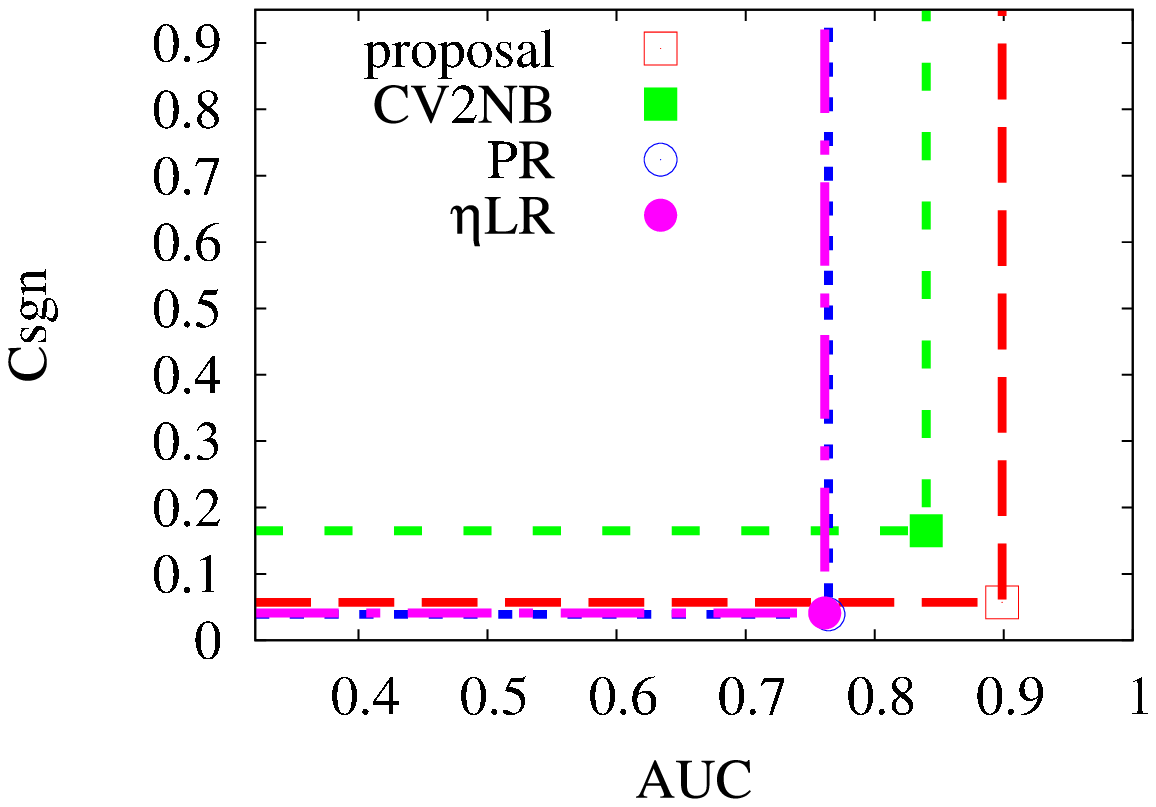}}
 \subfloat[Dutch Census]{\includegraphics[width=.32\textwidth]{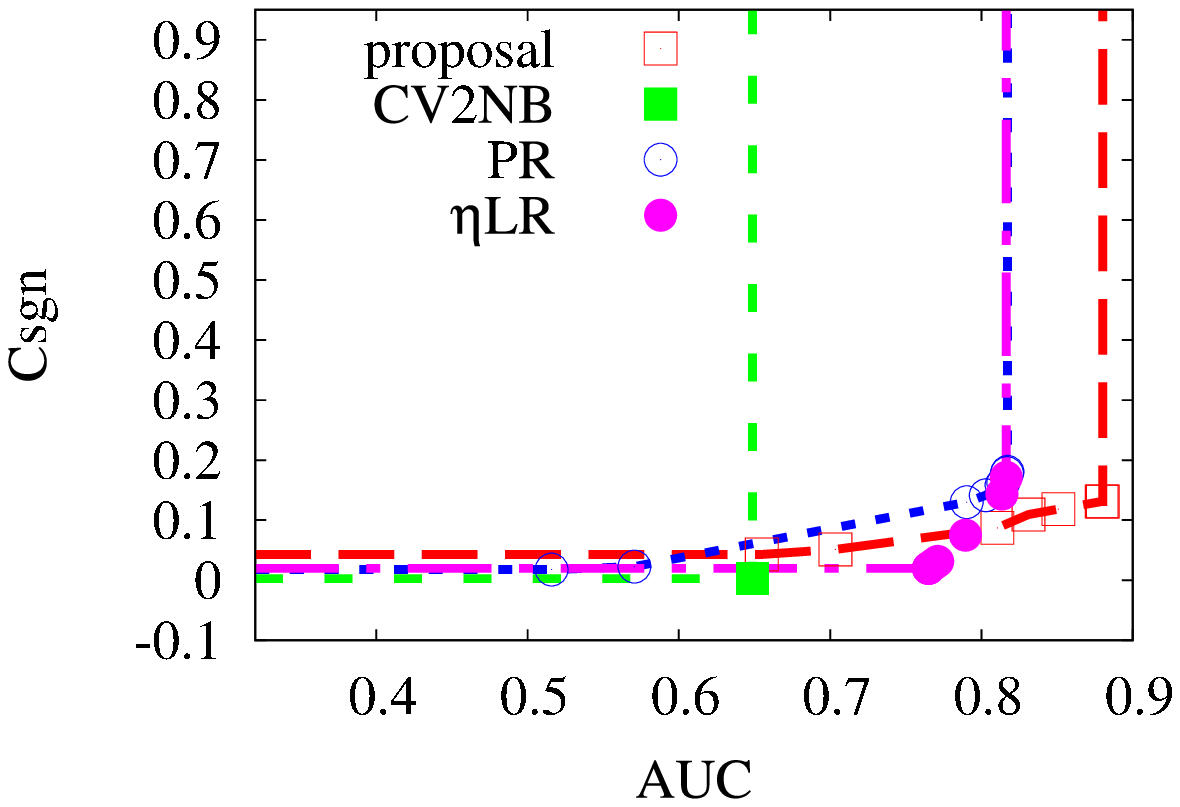}}\\
 \subfloat[Bank]{\includegraphics[width=.32\textwidth]{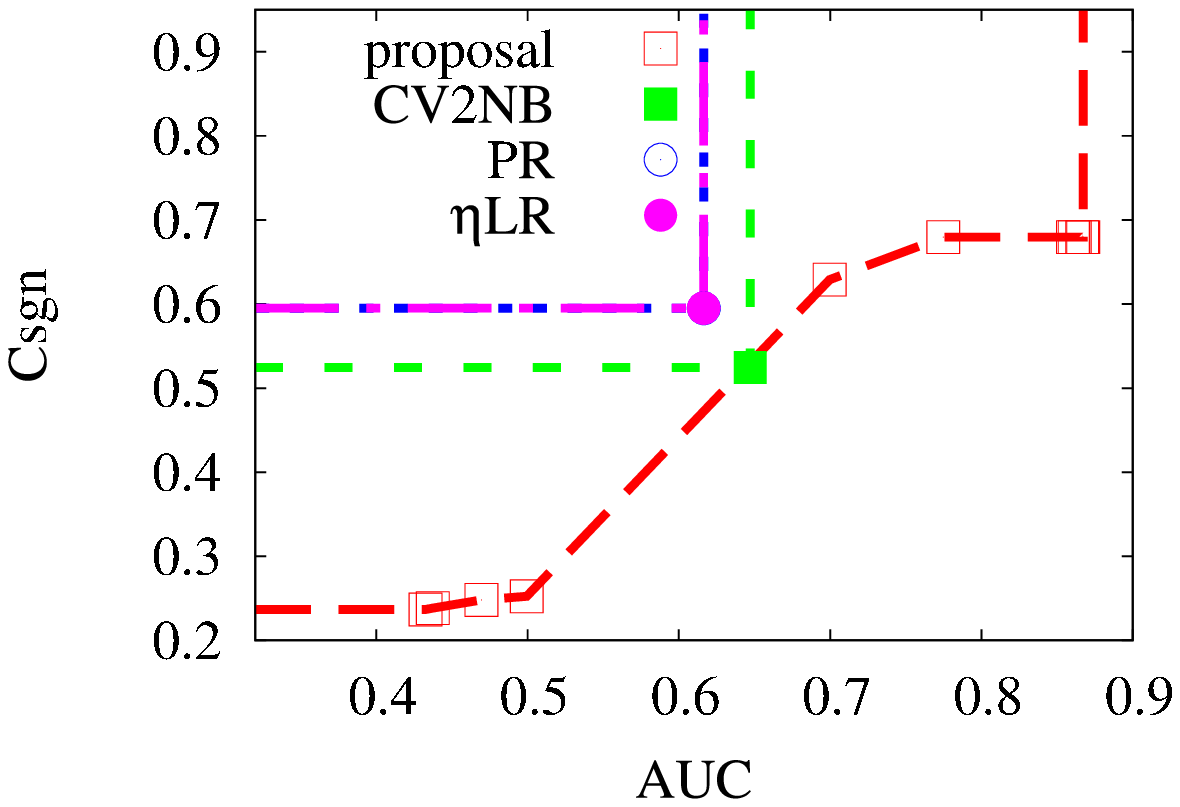}}
 \subfloat[German]{\includegraphics[width=.32\textwidth]{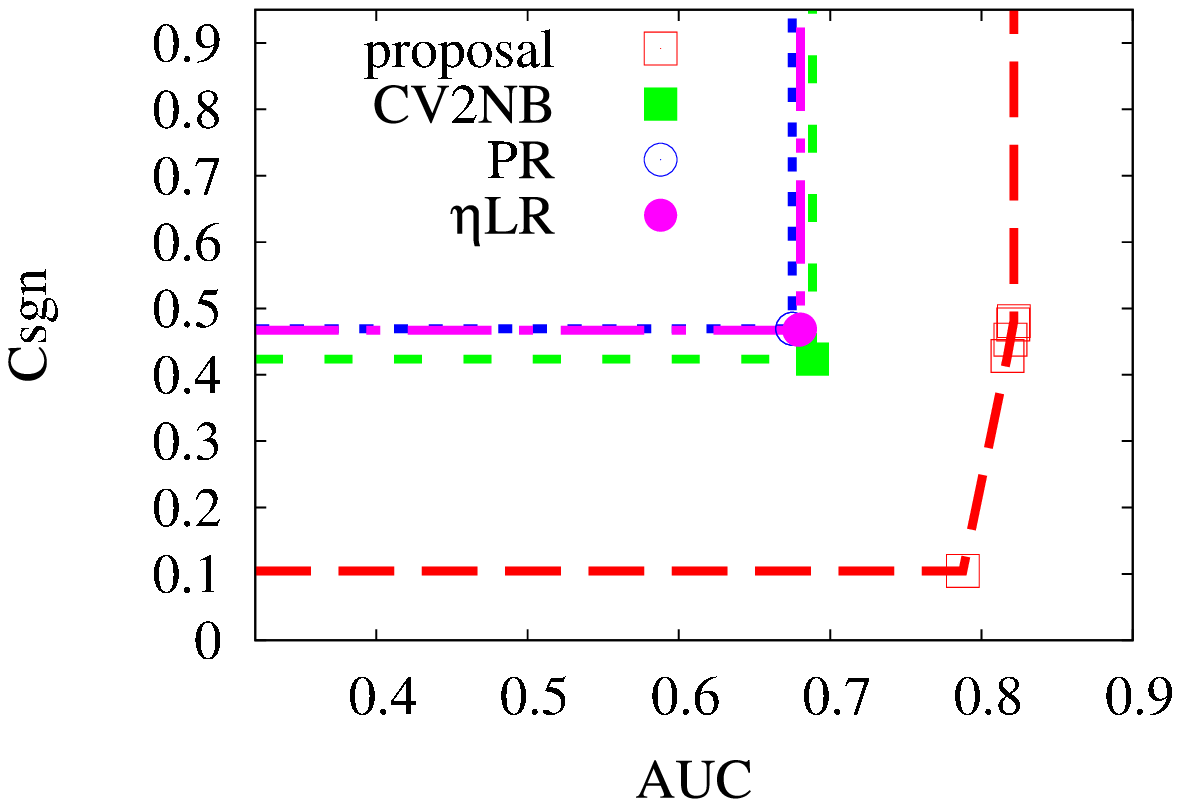}}
 \caption{Performance of CV2NB, PR, $\eta$LR, and neutral SVM (our proposal). The vertical axis shows the AUC, and horizontal axis shows $C_{n,sgn}(f,g)$. The points in these plots are omitted if they are dominated by others. The bottommost line shows limitations of neutralization performance, and the rightmost line shows limitations of classification performance, which are shown only as guidelines.}\label{fig:result-of-auc-nu}
\end{figure}

{\bf Results.}
\cref{fig:result-of-auc-nu} shows the classification performance (AUC) and neutralization performance ($C_{n,\rm sgn}(f,g)$) at different setting of neutralization parameter $\eta$. In the graph, the best result is shown at the right bottom. Since the classification performance and neutralization performance are in a trade-off relationship, as indicated by Theorem \cref{eq:empirical-risk-and-neutrality}, the results dominated by the other parameter settings are omitted in the plot for each method.

CV2NB achieves the best neutrality in Dutch Census, but is less neutral compared to the other methods in the rest of the datasets. In general, the classification performance of CV2NB is lower than those of the other methods due to the poor classification performance of na\'{i}ve Bayes. PR and $\eta$LR achieve competitive performance to neutral SVM in Adult and Dutch Census in term of the neutrality risk, but the results are dominated in term of AUC. Furthermore, the results of PR and $\eta$LR in Bank and German are dominated. The results of neutral SVM are dominant compared to the other methods in Bank and German dataset, and it is noteworthy that the neutral SVM achieves the best AUC in almost all datasets. This presumably reflects the superiority of SVM in the classification performance, compared to the na\'{i}ve Bayes and logistic regression.

\section{Conclusion}
We proposed a novel framework, NERM. NERM provides a framework that learns a target hypothesis that minimizes the empirical risk and that is empirically neutral in terms of risk to a given viewpoint hypothesis. Our contributions are as follows:
(1) We define NERM as a framework for guaranteeing the neutrality of classification problems. In contrast to existing methods, the NERM can be formulated as a convex optimization problem by using convex relaxation.
(2) We provide theoretical analysis of the generalization neutrality risk of NERM. The theoretical results show the approximation error of the generalization neutrality risk of NERM is uniformly upper-bounded by the Rademacher complexity of hypothesis class $g \fset{F}$ and $O(\sqrt{\ln(1/\delta)/n})$. Moreover, we derive a bound on the generalization neutrality risk for the optimal hypothesis corresponding to the neutralization parameter $\eta$.
(3) We present a specific learning algorithms for NERM, neutral SVM. We also extend the neutral SVM to the kernelized version.

Suppose the viewpoint is set to some private information.
Then, noting that neutralization reduces correlation between the target and viewpoint values, outputs obtained from the neutralized target hypothesis do not help to predict the viewpoint values.
Thus, neutralization realizes a certain type of privacy preservation.
In addition, as already mentioned, NERM can be interpreted as a variant of transfer learning by regarding the neutralization term as data-dependent prior knowledge.
Clarifying connection to privacy-preservation and transfer learning is remained as an area of future work.

\bibliographystyle{plainnat}
\bibliography{reference}

\begin{thebibliography}{10}
\providecommand{\url}[1]{\texttt{#1}}
\providecommand{\urlprefix}{URL }

\bibitem{bartlett2005}
Bartlett, P.L., Bousquet, O., Mendelson, S.: Local rademacher complexities. The
  Annals of Statistics  33(4),  1497--1537 (08 2005)

\bibitem{DBLP:journals/jmlr/BartlettM02}
Bartlett, P.L., Mendelson, S.: Rademacher and gaussian complexities: Risk
  bounds and structural results. Journal of Machine Learning Research  3,
  463--482 (2002)

\bibitem{DBLP:conf/icdm/CaldersKP09}
Calders, T., Kamiran, F., Pechenizkiy, M.: Building classifiers with
  independency constraints. In: Saygin, Y., Yu, J.X., Kargupta, H., Wang, W.,
  Ranka, S., Yu, P.S., Wu, X. (eds.) ICDM Workshops. pp. 13--18. IEEE Computer
  Society (2009)

\bibitem{Calders:2010:TNB:1842547.1842562}
Calders, T., Verwer, S.: Three naive bayes approaches for discrimination-free
  classification. Data Mining and Knowledge Discovery  21(2),  277--292 (Sep
  2010)

\bibitem{DBLP:journals/tist/ChangL11}
Chang, C.C., Lin, C.J.: Libsvm: A library for support vector machines. ACM TIST
   2(3), ~27 (2011)

\bibitem{DBLP:journals/jmlr/FanCL05}
Fan, R.E., Chen, P.H., Lin, C.J.: Working set selection using second order
  information for training support vector machines. Journal of Machine Learning
  Research  6,  1889--1918 (2005)

\bibitem{DBLP:conf/pkdd/FukuchiSK13}
Fukuchi, K., Sakuma, J., Kamishima, T.: Prediction with model-based neutrality.
  In: Blockeel, H., Kersting, K., Nijssen, S., Zelezn{\'y}, F. (eds.) ECML/PKDD
  (2). Lecture Notes in Computer Science, vol. 8189, pp. 499--514. Springer
  (2013)

\bibitem{DBLP:conf/nips/KakadeST08}
Kakade, S.M., Sridharan, K., Tewari, A.: On the complexity of linear
  prediction: Risk bounds, margin bounds, and regularization. In: Koller, D.,
  Schuurmans, D., Bengio, Y., Bottou, L. (eds.) NIPS. pp. 793--800. Curran
  Associates, Inc. (2008)

\bibitem{kamiran2010discrimination}
Kamiran, F., Calders, T., Pechenizkiy, M.: Discrimination aware decision tree
  learning. In: Data Mining (ICDM), 2010 IEEE 10th International Conference on.
  pp. 869--874. IEEE (2010)

\bibitem{kamishima.ecml.pkdd2012.fairness-aware}
Kamishima, T., Akaho, S., Asoh, H., Sakuma, J.: Fairness-aware classifier with
  prejudice remover regularizer. In: in Proceedings of the ECML/PKDD2012, Part
  II. vol. LNCS 7524, pp. 35--50. Springer (2012)

\bibitem{DBLP:journals/tkde/PanY10}
Pan, S.J., Yang, Q.: A survey on transfer learning. IEEE Trans. Knowl. Data
  Eng.  22(10),  1345--1359 (2010)

\bibitem{pariser2011filter}
Pariser, E.: The Filter Bubble: What The Internet Is Hiding From You. Viking,
  London (2011)

\bibitem{pedreschi2009measuring}
Pedreschi, D., Ruggieri, S., Turini, F.: Measuring discrimination in
  socially-sensitive decision records. In: Proceedings of the SIAM Int’l
  Conf. on Data Mining. pp. 499--514. Citeseer (2009)

\bibitem{resnick2011}
Resnick, P., Konstan, J., Jameson, A.: Measuring discrimination in
  socially-sensitive decision records. In: Proceedings of the 5th ACM
  Conference on Recommender Systems: Panel on The Filter Bubble. pp. 499--514
  (2011)

\bibitem{Shor:1985:MMN:3585}
Shor, N.Z., Kiwiel, K.C., Ruszcay\`{n}ski, A.: Minimization Methods for
  Non-differentiable Functions. Springer-Verlag New York, Inc., New York, NY,
  USA (1985)

\bibitem{vapnik1998statistical}
Vapnik, V.N.: Statistical learning theory  (1998)

\bibitem{DBLP:conf/icml/ZemelWSPD13}
Zemel, R.S., Wu, Y., Swersky, K., Pitassi, T., Dwork, C.: Learning fair
  representations. In: ICML (3). JMLR Proceedings, vol.~28, pp. 325--333.
  JMLR.org (2013)

\bibitem{zliobaite2011handling}
Zliobaite, I., Kamiran, F., Calders, T.: Handling conditional discrimination.
  In: Data Mining (ICDM), 2011 IEEE 11th International Conference on. pp.
  992--1001. IEEE (2011)

\end{thebibliography}

\appendix
\section{Proof of \cref{prop:equability-convert-to-max}}

\begin{proof}
Noting that
\begin{align}
 C^+_{\rm sgn}(f,g) + C^-_{\rm sgn}(f,g) = 1 \label{eq:trade-off-between-c}
\end{align}
holds, we have
 \begin{align}
  \abs{C^+_{\rm sgn}(f,g) - C^-_{\rm sgn}(f,g)} =& \abs{2C^+_{\rm sgn}(f,g) - 1}.
 \end{align}
Since $C^+_{\rm sgn}(f,g) \ge 1-\eta$ and $C^+_{\rm sgn}(f,g) \le \eta$ by the assumption,
 \begin{align}
 \abs{C^+_{\rm sgn}(f,g) - C^-_{\rm sgn}(f,g)} \le 2\eta -  1.
 \end{align}
\end{proof}

\section{Proof of \cref{thm:generalization-neutrality-bound}}

\begin{proof}

First, we derive the uniform bound on the relaxed convex empirical neutrality risk. For any $f \in \fset{F}$, we have
 \begin{align}
  C^\pm_\psi(f,g) \le C^\pm_{n,\psi}(f,g) + \sup_{f \in \fset{F}}(C^\pm_\psi(f,g) - C^\pm_{n,\psi}).
 \end{align}

Using the McDiarmid inequality, with probability $1-\delta/2$, we have
\begin{align}
 \sup_{f \in \fset{F}}(C^\pm_\psi(f,g) - C^\pm_{n,\psi}) \le
 \Mean[D_n]{\sup_{f \in \fset{F}}(C^\pm_\psi(f,g) - C^\pm_{n,\psi})} + c\sqrt{\frac{\ln(2/\delta)}{2n}}.
\end{align}

Application of the symmetrization technique yields the following bound:
\begin{align}
& \Mean[D_n]{\sup_{f \in \fset{F}}(C^\pm_\psi(f,g) - C^\pm_{n,\psi})} \\
=& \Mean[D_n]{\sup_{f \in \fset{F}}(\Mean[D'_n]{C^\pm_{n,\psi}(f,g)} - C^\pm_{n,\psi})} \label{eq:sym}\\
\le& \Mean[D_n,D'_n]{\sup_{h \in \psi \circ \pm g\fset{F}}\frac{1}{n}\sum_{i=1}^n(h(x_i) - h(x'_i))} \label{eq:jensen}\\
\le& L_\psi \Mean[D_n,D'_n]{\sup_{h \in \pm g\fset{F}}\frac{1}{n}\sum_{i=1}^n(h(x_i) - h(x'_i))} \label{eq:liphscitzf}\\
=& L_\psi \Mean[D_n,D'_n,\vec{\sigma}]{\sup_{h \in \pm g\fset{F}}\frac{1}{n}\sum_{i=1}^n\sigma_i(h(x_i) - h(x'_i))} \\
=& 2L_\psi\Mean[D_n,\vec{\sigma}]{\sup_{h \in \pm g\fset{F}}\frac{1}{n}\sum_{i=1}^n\sigma_i h(x_i)} \\
=& 2L_\psi\Mean[D_n,\vec{\sigma}]{\sup_{h \in g\fset{F}}\frac{1}{n}\sum_{i=1}^n\sigma_i h(x_i)} \\
=& 2L_\psi\mathcal{R}_n(g\fset{F})
\end{align}

The symmetrization technique (in the Glivenko--Cantelli theorem) was used to derive \cref{eq:sym}. The inequality in \cref{eq:jensen} is derived using the Jensen inequality and the convexity of $\sup(\cdot)$. The inequality in \cref{eq:liphscitzf} holds because $\psi(\cdot)$ is $L_\psi$-Lipschitz. Hence, with probability at least $1-\delta/2$,

\begin{align}
 C^\pm_\psi(f,g) \le C^\pm_{n,\psi}(f,g) + 2L_\psi\mathcal{R}_n(g\fset{F}) + c\sqrt{\frac{\ln(2/\delta)}{2n}} \label{eq:pm}
\end{align}

If $C^\pm_{n,\psi}(f,g) \le C^\mp_{n,\psi}(f,g)$ holds, we can show that the following bound holds with probability at least $1-\delta/2$ in a similar manner:

\begin{align}
 C^\pm_\psi(f,g) \le C^\mp_{n,\psi}(f,g) + 2L_\psi\mathcal{R}_n(g\fset{F}) + c\sqrt{\frac{\ln(2/\delta)}{2n}} \label{eq:mp}
\end{align}

Combining \cref{eq:pm} and \cref{eq:mp}, with probability at least $1-\delta$,

\begin{align}
  C_\psi(f,g)
 =& \max(C^+_\psi(f,g), C^-_\psi(f,g)) \\
 \le& C_{n,\psi}(f,g) + 2L_\psi\mathcal{R}_n(g\fset{F}) + c\sqrt{\frac{\ln(2/\delta)}{2n}}.
\end{align}
\end{proof}

\section{Proof of \cref{col:optimal-neutrality-bound}}

\begin{proof}
Based on the (A) conditions, the upper bound of the objective function of NERM with respect to $\hat{f}$ is given as follows:
 \begin{align}
   R_n(\hat{f}) + \Omega(\hat{f}) + \eta C_{n,\psi}(\hat{f},g)
    \le& R_n(f_0) + \Omega(f_0) + \eta C_{n,\psi}(f_0,g) \\
  =& \phi(0) + \eta\psi(0) .
 \end{align}

Since $R_n(f) \ge 0$ and $\Omega(f) \ge 0$, we have
 \begin{align}
  \eta C_{n,\psi}(\hat{f},g) \le& \phi(0) + \eta\psi(0) \\
  C_{n,\psi}(\hat{f},g) \le& \psi(0) + \phi(0)\frac{1}{\eta} .
 \end{align}
\end{proof}

\section{Optimization of Primal Neutral SVM}
Neutral SVM is formulated as the following optimization problem
\begin{align}
 \min_{\vec{w} \in \RealSet^d,b \in \RealSet} ~ & \Psi(\vec{w},b) = \sum_{i=1}^n \max(0,1-y_i (\vec{w}^T\vec{x}_i + b)) + \frac{1}{2} \norm{\vec{w}}_2^2 + \eta C_{n,\psi}(\vec{w},b,g) \label{eq:optimization-primal}
\end{align}
where
\begin{align}
 & C_{n,\psi}(\vec{w},b,g) = \max(C_{n,\psi}^+(\vec{w},b,g), C_{n,\psi}^-(\vec{w},b,g)), \\
 & C_{n,\psi}^\pm(\vec{w},b,g) = \sum_{i=1}^n \max(0, 1 \mp g(\vec{x}_i)(\vec{w}^T\vec{x}_i + b)) .
\end{align}
Since the problem of \cref{eq:optimization-primal} can be solved by applying the subgradient method \cite{Shor:1985:MMN:3585}, we provide the subgradient of the objective function of \cref{eq:optimization-primal}.

For convenience, we introduce the following notations.
For a set $\set{C} \subseteq \RealSet^d \times \RealSet$ and constant $a \in \RealSet$, we denote $a\set{C} = \{ (a\vec{w}, ab) | (\vec{w}, b) \in \set{C}\}$.
For any sets $\set{C}_1 \subseteq \RealSet^d \times \RealSet$ and $\set{C}_2 \subseteq \RealSet^d \times \RealSet$, let $\set{C}_1 + \set{C}_2 = \{ (\vec{w}_1 + \vec{w}_2, b_1 + b_2) | (\vec{w}_1, b_1) \in \set{C}_1, (\vec{w}_2, b_2) \in \set{C}_2 \}$, and let ${\bf Co}~ \set{C}_1 \cup \set{C}_2$ be convex hull of $\set{C}_1$ and $\set{C}_2$, i.e., ${\bf Co}~ \set{C}_1 \cup \set{C}_2 = \{ \alpha c_1 + (1-\alpha)c_2 | c_1 \in \set{C}_1, c_2 \in \set{C}_2, \alpha \in [0,1] \}$.

The subgradient of the objective function of \cref{eq:optimization-primal} is derived as follows:

\begin{align}
 \partial \Psi(\vec{w},b) =& \sum_{i=1}^n \partial\ell(y_i, \vec{w}^T\vec{x}_i + b) + \lambda\{ (\vec{w}, 0) \} + \eta \partial C_{n,\psi}(\vec{w},b,g)  \label{eq:subgradient-primal}
\end{align}
where
\begin{align}
  \partial \ell(t_i,\vec{w}^T\vec{x}_i + b) =& \begin{dcases}
    \{(-t_i\vec{x}_i, 1) \} & \mbox{if } \vec{w}^T\vec{x}_i + b < 1, \\
    \{(\vec{0}, 0)       \} & \mbox{if } \vec{w}^T\vec{x}_i + b > 1, \\
    \{(-\alpha t_i\vec{x}_i, \alpha) | \alpha \in [0,1] \} & \mbox{if } \vec{w}^T\vec{x}_i + b = 1 ,
  \end{dcases} \label{eq:subgradient-hinge}
\end{align}
\begin{align}
  & \partial C_{n,\psi}(\vec{w},b,g) = \\
  & \begin{dcases}
    \partial C_{n,\psi}^+(\vec{w},b,g) & \mbox{if } C_{n,\psi}^+(\vec{w},b,g) > C_{n,\psi}^-(\vec{w},b,g),\\
    \partial C_{n,\psi}^-(\vec{w},b,g) & \mbox{if } C_{n,\psi}^+(\vec{w},b,g) < C_{n,\psi}^-(\vec{w},b,g),\\
    {\bf Co}( \partial C_{n,\psi}^+(\vec{w},b,g) \cup \partial C_{n,\psi}^-(\vec{w},b,g)) & \mbox{if } C_{n,\psi}^+(\vec{w},b,g) = C_{n,\psi}^-(\vec{w},b,g), \\
  \end{dcases} \label{eq:subgradient-cpsi}
\end{align}
\begin{align}
  \partial C_{n,\psi}^\pm(\vec{w},b,g) =& \sum_{i=1}^n \partial\ell(\pm v_i, \vec{w}^T\vec{x}_i + b) .\label{eq:subgradient-cpsi-pm}
\end{align}
\cref{eq:subgradient-hinge}, \cref{eq:subgradient-cpsi} and \cref{eq:subgradient-cpsi-pm} denote the subgradient of hinge loss function, $C_{n,\psi}(\vec{w},b,g)$ and $C_{n,\psi}^\pm(\vec{w},b,g)$, respectively.

\section{Kernelization of Neutral SVM}
In this section, we derive the dual problem. From the dual, kernelization of neutral SVM is readily obtained.
First, introducing slack variables $\vec{\xi}$, $\vec{\xi}^\pm$, and $\zeta$ into \cref{eq:optimization-primal}, we have
\begin{align}
 \min_{\vec{w}, b, \vec{\xi}, \vec{\xi}^\pm, \zeta} ~ & \sum_{i=1}^n \xi_i + \frac{\lambda}{2}\norm{\vec{w}}_2^2 + \eta \zeta \label{eq:optimization-constrants}\\
 \mbox{\rm sub to} ~ & \sum_{i=1}^n \xi^+_i \le \zeta,
                     , \sum_{i=1}^n \xi^-_i \le \zeta, \\
                     & 1 - y_i(\vec{w}^T\vec{x}_i + b) \le \xi_i, \\
                     & 1 - v_i(\vec{w}^T\vec{x}_i + b) \le \xi^+_i, \\
                     & 1 + v_i(\vec{w}^T\vec{x}_i + b) \le \xi^-_i, \\
                     & \xi_i \ge 0, \xi^+_i \ge 0, \xi^-_i \ge 0, \zeta \ge 0
\end{align}
where $v_i = g(\vec{x}_i)$.
From the Lagrange relaxation of the primal problem \cref{eq:optimization-constrants}, we have the Lagrange relaxation function as
\begin{align}
  & L(\vec{w}, b, \vec{\xi}, \vec{\xi}^\pm, \zeta, \vec{\alpha}, \vec{\beta}^\pm, \nu^\pm, \vec{\delta}) \\
 =& \sum_{i=1}^n \xi_i + \nu^+(\sum_{i=1}^n \xi^+_i - \zeta) + \nu^-(\sum_{i=1}^n \xi^-_i - \zeta) + \frac{\lambda}{2}\norm{\vec{w}}_2^2 + \eta\zeta \\
  & + \sum_{i=1}^n\alpha_i(1-y_i(\vec{w}^T\vec{x}_i + b) - \xi_i) \\
  & + \sum_{i=1}^n\beta^+_i(1-v_i(\vec{w}^T\vec{x}_i + b) - \xi^+_i) \\
  & + \sum_{i=1}^n\beta^-_i(1+v_i(\vec{w}^T\vec{x}_i + b) - \xi^-_i) \\
  & - \sum_{i=1}^n (\delta_{\xi, i}\xi_i + \delta_{\xi, +, i}\xi^+_i + \delta_{\xi, -, i}\xi^-_i) - \delta_{\zeta}\zeta .
\end{align}
By partial differentiating $L$ for each variables, we have
\begin{align}
 \partial_{\vec{w}}L =& \lambda\vec{w} - \sum_{i=1}^n(\alpha_iy_i + \beta^+_iv_i - \beta^-_iv_i)\vec{x}_i, \\
 \partial_{b}L =& -\sum_{i=1}^n(\alpha_iy_i + \beta^+_iv_i - \beta^-_iv_i), \\
 \partial_{\xi_i}L =& 1 - \alpha_i - \delta_{\xi, i}, \\
 \partial_{\xi^+_i}L =& \nu^+ - \beta^+_i - \delta_{\xi, +, i}, \\
 \partial_{\xi^-_i}L =& \nu^- - \beta^-_i - \delta_{\xi, -, i}, \\
 \partial_{\zeta}L =& \eta - \nu^+ - \nu^- - \delta_{\zeta} .
\end{align}
Letting these partial differentiations be zero, we have the dual problem as
\begin{align}
 \max_{\vec{\alpha}, \vec{\beta}^\pm} ~ & \lambda\sum_{i=1}^nb_i - \frac{1}{2}\sum_i^n\sum_j^n a_ia_j \vec{x}_i^T\vec{x}_j \\
 \mbox{\rm sub to} ~ & \sum_i^n a_i = 0, \\
                     & 0 \le \alpha_i \le 1, 0 \le \beta^+_i, 0 \le \beta^-_i, \beta^+_i + \beta^-_i \le \eta
\end{align}
where $b_i = \alpha_i+\beta^+_i+\beta^-_i$ and $a_i = \alpha_iy_i + \beta^+_iv_i - \beta^-_iv_i$. By replacing the inner product $\vec{x}_i^T\vec{x}_j$ for the kernel function $\vec{x}_i^T\vec{x}_j = k(x_i, x_j)$, kernelization of neutral SVM is obtained as
\begin{align}
 \max_{\vec{\alpha}, \vec{\beta}^\pm} ~ & \lambda\sum_{i=1}^nb_i - \frac{1}{2}\sum_i^n\sum_j^n a_i a_i k(x_i,x_j) \label{eq:optimization-kernel}\\
 \mbox{\rm sub to} ~ & \sum_i^n a_i = 0, \\
                     & 0 \le \alpha_i \le 1, 0 \le \beta^+_i, 0 \le \beta^-_i, \beta^+_i + \beta^-_i \le \eta .
\end{align}

\section{SMO-like Optimization for kernelized neutral SVM}
The optimization of \cref{eq:optimization-kernel} is an instance of {\em Quadratic Programming (QP)} and it can be solved by general QP solvers, whereas it does not scale well with large samples due to its large memory consumption.
We also show that the application of the well known {\em Sequential Minimal Optimization (SMO)} technique to our neutral SVM.
In order to reduce memory consumption, the SMO modifies only a subset of the parameters per iteration.
The well-known SVM solver LIBSVM \cite{DBLP:journals/tist/ChangL11} solves the dual problem of the original SVM by using SMO.
In this section, we apply the SMO technique to our neutral SVM in the same manner of LIBSVM.

We denote the gram matrix $\mat{K}$ where $K_{ij} = k(x_i,x_j)$ and let
\begin{align}
 \vec{\gamma} =& (\begin{array}{ccc} \vec{\alpha}^T & \vec{\beta}^{+T} & \vec{\beta}^{-T} \end{array})^T ,\\
 \vec{t} =& (\begin{array}{ccccccccc} y_1 & ... & y_n & v_1 & ... & v_n & -v_1 & ... & v_n \end{array})^T ,\\
 \mat{\Sigma} =& \left(\begin{array}{ccccccccc}
  y_1     &        & \mat{0} & v_1     &        & \mat{0} & -v_1    &        & \mat{0} \\
          & \ddots &         &         & \ddots &         &         & \ddots &         \\
  \mat{0} &        & y_n     & \mat{0} &        & v_n     & \mat{0} &        & -v_n
 \end{array}\right)^T , \\
 \mat{Q} =& \mat{\Sigma}\mat{K}\mat{\Sigma}^T .
\end{align}
Then, the optimization problem of \cref{eq:optimization-kernel} can be rearranged as
\begin{align}
 \min_{\vec\gamma} ~ & \frac{1}{2}\vec\gamma^T \mat{Q} \vec\gamma - \lambda \vec{1}^T\vec\gamma \\
 \mbox{\rm sub to} ~ & \vec{t}^T\vec\gamma = 0, \vec\gamma \ge 0,\\
                     & \gamma_i \le 1 ~\forall i=1,...,n,\\
                     & \gamma_i + \gamma_{i+n} \le \eta ~\forall i=n+1,...,2n.
\end{align}
\cref{alg:smo} shows the SMO-like algorithm for our neutral SVM, and the details of working set selection in step 3 and parameter update in step 4 is described in \cref{sec:opt-selection} and \cref{sec:opt-update}, respectively.
The stopping criteria in step 6 is described in \cref{sec:opt-convergence}.

\begin{algorithm}[t]
 \caption{Optimize neutral SVM with SMO-like algorithm}\label{alg:smo}
 \begin{algorithmic}[1]
  \STATE Find $\vec\gamma^1$ as the initial feasible solution. Set $k=1$
  \REPEAT
   \STATE Select Working Set $B = \{i, j\} \subset \{1,...,3n\} ~ (i \ne j)$ \hfill \cref{sec:opt-selection}
   \STATE Update $\vec\gamma^k$ to $\vec\gamma^{k+1}$ \hfill \cref{sec:opt-update}
   \STATE $k \leftarrow k+1$
  \UNTIL{Convergence ~ (\cref{sec:opt-convergence})}
 \end{algorithmic}
\end{algorithm}

\subsection{Update Parameters}\label{sec:opt-update}
In this subsection, we provide the procedure for update of the parameters.
We focus on two parameters $\gamma_i$ and $\gamma_j$.
Suppose we update these parameters by $\gamma'_i = \gamma_i + \delta_i$ and $\gamma'_j = \gamma_j + \delta_j$.

The sub-problem of \cref{eq:optimization-kernel} with the two variables $\delta_i$ and $\delta_j$ is as follows:
\begin{align}
 \min_{\delta_i, \delta_j} ~ & \frac{1}{2}(Q_{ii}\delta_i^2 + 2Q_{ij}\delta_i\delta_j + Q_{jj}\delta_j^2)  \\
  &+ \delta_i\sum_{\ell=1}^n Q_{i\ell}\gamma_\ell + \delta_j\sum_{\ell=1}^n Q_{j\ell}\gamma_\ell - \lambda\delta_i - \lambda\delta_j + \mbox{const}. \label{eq:selected-objective}
\end{align}
We do not consider the inequality constraints for now.
From the equality constraint, we have
\begin{align}
 t_i \delta_i + t_j \delta_j = 0. \label{eq:constraints-linear-delta}
\end{align}

By substituting \cref{eq:constraints-linear-delta} into \cref{eq:selected-objective}, the sub-problem of \cref{eq:optimization-kernel} with respect to $\gamma_i$ and $\gamma_j$ is reduced to the optimization problem of the single variable $\delta_i$:
\begin{align}
 \min_{\delta_i} ~ & \frac{1}{2}(Q_{ii} - 2 t_i t_j^{-1} Q_{ij} + t_i^2 t_j^{-2} Q_{jj})\delta_i^2  \\&
  - (\lambda - t_it_j^{-1}\lambda - \sum_{\ell=1}^n Q_{i\ell}\gamma_\ell + t_i t_j^{-1} \sum_{\ell=1}^n Q_{j\ell}\gamma_\ell)\delta_i + \mbox{const}. \label{eq:optimization-one-variable}
\end{align}

Since  \cref{eq:optimization-one-variable} has the closed from, the optimal solution is analytically obtained as
\begin{align}
 \delta_i =& \frac{\lambda - \tau_{ij}\lambda - \sum_{\ell=1}^n Q_{i\ell}\gamma_\ell + \tau_{ij}\sum_{\ell=1}^n Q_{j\ell}\gamma_\ell}{Q_{ii} - 2\tau_{ij}Q_{ij} + \tau_{ij}^2Q_{jj}}
\end{align}
where $\tau_{ij} = t_it_j^{-1}$.
The optimal solution with respect to $\delta_j$ can be readily derived using \cref{eq:constraints-linear-delta}.

Next, we consider the feasible region of $\gamma_i$.  The feasible region of $\gamma_i$ is determined by the inequality constraints in \cref{eq:optimization-kernel}.
Two different feasible regions can be determined depending on the selection of $\gamma_i$ and $\gamma_j$.





{\bf Case 1. $i>n, j=i+n$.}

In this case, $\gamma_i$ and $\gamma_j$ is selected as the $i$-th element in $\vec{\beta}^+$ and $\vec{\beta}^-$, respectively.
The feasible region of $\gamma'_i$ is specified by the inequality constraints in \cref{eq:optimization-kernel}, $\gamma'_i \ge 0$, $\gamma'_j \ge 0$ and $\gamma'_i + \gamma'_j \le \eta$ as follows:
\begin{align}
 \gamma'_i \in \begin{cases}
   [\max(0, a^{11}_{ij}), a^{12}_{ij} ] & \mbox{if } \tau_{ij} \ge 1 , \\
   [0, \min(a^{11}_{ij}, a^{12}_{ij}) ] & \mbox{if } 0 \le \tau_{ij} < 1 , \\
   [\max(0, a^{12}_{ij}), a^{11}_{ij} ] & otherwise  \\
 \end{cases}
\end{align}
where
\begin{align}
 a^{11}_{ij} =& (1-\tau_{ij}^{-1})^{-1}\gamma_i + (\tau_{ij}-1)^{-1}\gamma_j - (\tau_{ij}-1)^{-1}\eta, \mbox{ and} \\
 a^{12}_{ij} =& \gamma_i + \tau_{ij}^{-1}\gamma_j.
\end{align}

The feasible region of $\gamma'_j$ is readily derived from above and \cref{eq:constraints-linear-delta}.

{\bf Case 2. }

All selections of $i$ and $j$ that are not contained in Case 1 belong to Case 2.
The feasible region of $\gamma'_i$ and $\gamma'_j$ specified by the inequality constraints in \cref{eq:optimization-kernel} is $\gamma'_i \in [0, u_i]$ and $\gamma'_j \in [0, u_j]$, respectively, where
\begin{align}
 u_i =& \begin{cases}
   1 & \mbox{if } 1 \le i \le n ~ (\mbox{$\gamma_i$ is $\alpha_i$}), \\
   \eta - \gamma_{i+n} & \mbox{if } n < i \le 2n ~ (\mbox{$\gamma_i$ is $\beta^+_{i-n}$}), \\
   \eta - \gamma_{i-n} & \mbox{if } 2n < i \le 3n ~ (\mbox{$\gamma_i$ is $\beta^-_{i-2n}$}) .
 \end{cases}
\end{align}
Thus, the feasible region of $\gamma'_i$ can be derived, as follows:
\begin{align}
 \gamma'_i \in \begin{cases}
   [\max(0, a^{21}_{ij}), \min(u_i, a^{22}_{ij})] & \mbox{if } \tau_{ij} \ge 0, \\
   [\max(0, a^{22}_{ij}), \min(u_i, a^{21}_{ij})] & otherwise
 \end{cases}
\end{align}
where
\begin{align}
 a^{21}_{ij} =& \gamma_i + \tau_{ij}^{-1}\gamma_j - \tau_{ij}^{-1}u_j, \mbox{ and} \\
 a^{22}_{ij} =& \gamma_i + \tau_{ij}^{-1}\gamma_j.
\end{align}

The feasible region of $\gamma'_j$ can be derived in the same manner.

\subsection{Stopping Criteria}\label{sec:opt-convergence}

The stopping criteria of our kernelized neutral SVM can be derived in the same manner as LIBSVM \cite{DBLP:journals/tist/ChangL11}.
The Karush-Kuhn-Tucker (KKT) optimally condition of problem \cref{eq:optimization-kernel} implies that a feasible $\vec\gamma$ is a stationary point of \cref{eq:optimization-kernel} if and only if there exists a number $p$ and two nonnegative vector $\vec\zeta$ and $\vec\xi$ such that,
\begin{gather}
 \nabla f(\vec{\gamma}) + p \vec{t} = \vec{\zeta} - \vec{\xi}, \\
 \lambda_i \gamma_i = 0, \lambda_i \ge 0, \xi_i \ge 0, i = 1,...,3n, \\
 \xi_i (1 - \gamma_i) = 0, i = 1,...,n, \\
 \xi_i (\eta - \gamma_i - \gamma_{i+n}) = 0, \xi_i = \xi_{i+n}, i = n+1,...,2n
\end{gather}
where $\nabla f(\vec{\gamma}) = \mat{Q}\vec{\gamma} + \lambda\vec{1}$.
These conditions can be rewritten as
\begin{align}
 \nabla_i f(\vec{\gamma}) + p t_i
 \begin{dcases}
   \ge 0 & \mbox{if } 1 \le i \le n, \gamma_i < 1 \\
         & \mbox{or } n < i \le 2n, \gamma_i + \gamma_{i+n} < \eta \\
         & \mbox{or } 2n < i \le 3n, \gamma_{i-n} + \gamma_{i} < \eta, \\
   \le 0 & \mbox{if } \gamma_i > 0 .
 \end{dcases}
\end{align}
These conditions are equivalent to that there exists $p$ such that
\begin{align}
 m(\vec{\gamma}) \le p \le M(\vec{\gamma})
\end{align}
where
\begin{align}
 m(\vec{\gamma}) =& \max_{i \in I_{\rm up}(\vec{\gamma})} -t_i^{-1}\nabla_if(\vec{\gamma}), \\
 M(\vec{\gamma}) =& \min_{i \in I_{\rm low}(\vec{\gamma})} -t_i^{-1}\nabla_if(\vec{\gamma})
\end{align}
and
\begin{align}
 I_{\rm up}(\vec{\gamma}) \coloneqq \{ i |  & 1 \le i \le n, \gamma_i < 1, t_i \ge 0 \mbox{ or } \\
                                            & n < i \le 2n, \gamma_i < \eta - \gamma_{i+n}, t_i \ge 0 \mbox{ or }\\
                                            & 2n < i \le 3n, \gamma_i < \eta - \gamma_{i-n}, t_i \ge 0 \mbox{ or } \\
                                            & \gamma_i > 0, t_i \le 0 \}, \\
 I_{\rm low}(\vec{\gamma}) \coloneqq \{ i | & 1 \le i \le n, \gamma_i < 1, t_i \le 0 \mbox{ or } \\
                                            & n < i \le 2n, \gamma_i < \eta - \gamma_{i+n}, t_i \le 0 \mbox{ or }\\
                                            & 2n < i \le 3n, \gamma_i < \eta - \gamma_{i-n}, t_i \le 0 \mbox{ or } \\
                                            & \gamma_i > 0, t_i \ge 0 \}. \\
\end{align}
That is, a feasible $\vec\gamma$ is a stationary point of problem \cref{eq:optimization-kernel} if and only if
\begin{align}
 m(\vec{\gamma}) \le M(\vec{\gamma}).
\end{align}
Thus, the stopping condition of \cref{alg:smo} is
\begin{align}
 m(\vec{\gamma}^k) - M(\vec{\gamma}^k) \le \epsilon
\end{align}
where $\epsilon$ is a tolerance.

\subsection{Working Set Selection}\label{sec:opt-selection}
We introduce the working set selection of kernelized neutral SVM in the same manner as WSS2 \cite{DBLP:journals/jmlr/FanCL05}.
We provide the following WSS2 of kernelized neutral SVM : \\
{\bf WSS2 of kernelized neutral SVM}
\begin{itemize}
 \item Select
 \begin{align}
  i \in& \argmax_i \{ -t_\ell^{-1}\nabla_if(\vec{\gamma}) | \ell \in I_{\rm up}(\vec{\gamma}) \} , \mbox{ and then} \\
  j \in& \argmin_\ell \{ -\frac{B^{2}_{i\ell}}{A_{i\ell}} | \ell \in I_{\rm low}(\vec{\gamma}), -t_i^{-1}\Delta_\ell f(\vec{\gamma}) < -t_i^{-1}\nabla_if(\vec{\gamma}) \}.
 \end{align}
 where
 \begin{align}
  A_{ij} =& Q_{ii} - 2\tau_{ij}Q_{ij} + \tau_{ij}^2Q_{jj}, \mbox{ and} \\
  B_{ij} =& \lambda - \tau_{ij}\lambda - \sum_{\ell=1}^n Q_{i\ell}\gamma_\ell + \tau_{ij}\sum_{\ell=1}^n Q_{j\ell}\gamma_\ell.
 \end{align}
 \item Return $B = \{i,j\}$.
\end{itemize}
The selection with respect to $j$ is derived from \cref{eq:selected-objective} in the same manner as \cite{DBLP:journals/jmlr/FanCL05}.

\end{document}